\documentclass[10pt,twocolumn,letterpaper]{article}

\usepackage{iccv}
\usepackage{times}
\usepackage{epsfig}
\usepackage{graphicx}
\usepackage{amsmath}
\usepackage{amssymb}

\usepackage{cuted}
\usepackage{capt-of}
\usepackage{comment}
\usepackage{booktabs}
\usepackage{lipsum}
\newcommand{\bftab}{\fontseries{b}\selectfont}

\usepackage[pagebackref=true,breaklinks=true,letterpaper=true,colorlinks,bookmarks=false]{hyperref}

\iccvfinalcopy %

\newcommand{\myparagraph}[1]{\vspace{.2cm} \noindent \textbf{#1} \:}

\usepackage[american]{babel}
\addto\extrasamerican{%
}

\begin{document}

\title{AttrLostGAN: Attribute Controlled Image Synthesis \\ from Reconfigurable Layout and Style}

\author{Stanislav Frolov\textsuperscript{1,2}, Avneesh Sharma\textsuperscript{1}, J\"orn Hees\textsuperscript{2}, Tushar Karayil\textsuperscript{2}, Federico Raue\textsuperscript{2}, Andreas Dengel\textsuperscript{1,2}\\
\textsuperscript{1}Technical University of Kaiserslautern \\
\textsuperscript{2}German Research Center for Artificial Intelligence (DFKI)\\
{\tt\small firstname.lastname@dfki.de, asharma@rhrk.uni-kl.de
}
\\[-3.0ex]
}

\maketitle

\begin{strip}
\centering
\includegraphics[width=\textwidth]{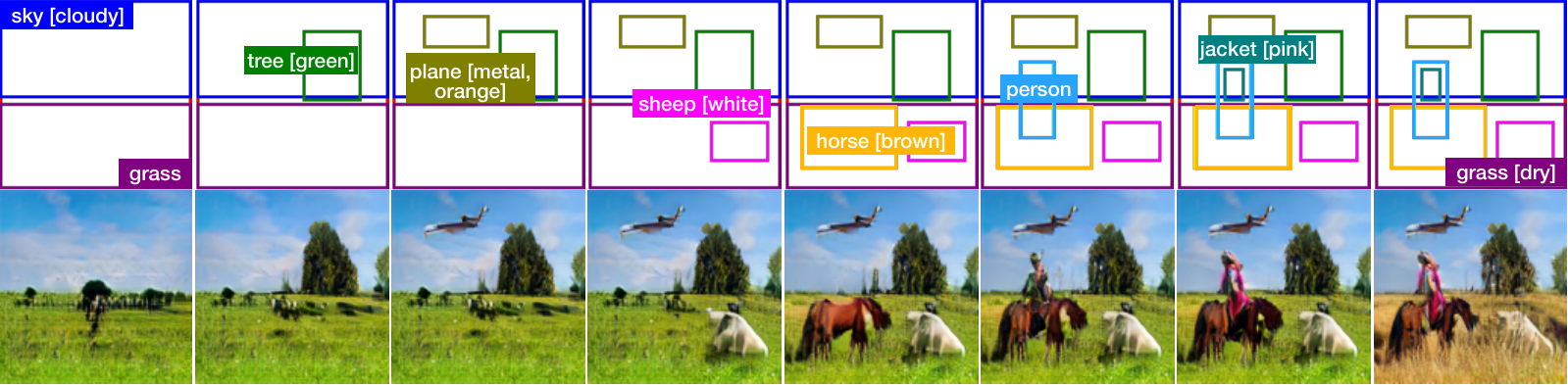}
\captionof{figure}{Generated images using a reconfigurable layout and atttributes to control the appearance of individual objects.
   \textit{From left to right:} add tree [green], add plane [metal, orange], add sheep [white], add horse [brown], add person, add jacket [pink], grass $\rightarrow$ grass [dry].
\label{fig:iterative}}
\end{strip}

\begin{abstract}
Conditional image synthesis from layout has recently attracted much interest.
Previous approaches condition the generator on object locations as well as class labels but lack fine-grained control over the diverse appearance aspects of individual objects.
Gaining control over the image generation process is fundamental to build practical applications with a user-friendly interface.
In this paper, we propose a method for attribute controlled image synthesis from layout which allows to specify the appearance of individual objects without affecting the rest of the image.
We extend a state-of-the-art approach for layout-to-image generation to additionally condition individual objects on attributes.
We create and experiment on a synthetic, as well as the challenging Visual Genome dataset.
Our qualitative and quantitative results show that our method can successfully control the fine-grained details of individual objects when modelling complex scenes with multiple objects.
Source code, dataset and pre-trained models are publicly available \footnote{https://github.com/stanifrolov/AttrLostGAN}.
\end{abstract}

\section{Introduction}
\label{sec:intro}
The advent of Generative Adversarial Networks (GANs) \cite{GAN} had a huge influence on the progress of image synthesis research and applications.
Starting from low-resolution, gray-scale face images, current methods can generate high-resolution face images which are very difficult to distinguish from real photographs \cite{StyleGAN}.
While unconditional image synthesis is interesting, most practical applications require an interface which allows users to specify what the model should generate.
In recent years, conditional generative approaches have used class labels \cite{cGAN,BigGAN}, images \cite{Isola_2017_CVPR,Zhu_2017_ICCV}, text \cite{Reed2016,StackGAN,AttnGAN}, speech \cite{wang2020domain,Choi2020FromIT}, layout \cite{Layout2Im,LostGAN}, segmentation masks \cite{Wang2017HighResolutionIS,Dong2017}, or combinations of them \cite{InferGAN}, to gain control over the image generation process.
However, most of these approaches are ``one-shot'' image generators which do not allow to reconfigure certain aspects of the generated image.

While there has been much progress on iterative image manipulation, researchers have so far not investigated how to gain better control over the image generation process of complex scenes with multiple interacting objects.
To allow the user to create a scene that reflects what he/she has in mind, the system needs to be capable of iteratively and interactively updating the image.
A recent approach by Sun and Wu \cite{LostGAN} takes a major step towards this goal by enabling reconfigurable spatial layout and object styles.
In their method, each object has an associated latent style code (sampled from a normal distribution) to create new images.
However, this implies that users do not have true control over the specific appearance of objects.
This lack of control also translates into the inability to specify a style (i.e., to change the color of a shirt from red to blue one would need to sample new latent codes and manually inspect whether the generated style conforms to the requirement).
Being able to not just generate, but control individual aspects of the generated image without affecting other areas is vital to enable users to generate what they have in mind.
To overcome this gap and give users control over style attributes of individual objects, we propose to extend their method to additionally incorporate attribute information.
To that end, we propose Attr-ISLA as an extension of the Instance-Sensitive and Layout-Aware feature Normalization (ISLA-Norm) \cite{LostGAN} and use an adversarial hinge loss on object-attribute features to encourage the generator to produce objects reflecting the input attributes.
At inference time, a user can not only reconfigure the location and class of individual objects, but also specify a set of attributes.
See \autoref{fig:iterative} for an example of reconfigurable layout-to-image generation guided by attributes using our method.
Since we continue to use latent codes for each object and the overall image, we can generate diverse images of objects with specific attributes.
This approach not only drastically improves the flexibility but also allows the user to easily articulate the contents of his mind into the image generation process. Our contributions can be summarized as following:

\begin{itemize}
    \item we propose a new method called AttrLostGAN, which allows attribute controlled image generation from reconfigurable layout;
    \item we extend ISLA to Attr-ISLA thereby gaining additional control over attributes;
    \item we create and experiment on a synthetic dataset to empirically demonstrate the effectiveness of our approach;
    \item we evaluate our model on the challenging Visual Genome dataset both qualitatively and quantitatively and achieve state-of-the-art performance.
\end{itemize}

\section{Related Work}
\label{sec:relatedwork}

\myparagraph{Class-Conditional Image Generation}
Generating images given a class label is arguably the most direct way to gain control over what image to generate.
Initial approaches concatenate the noise vector with the encoded label to condition the generator \cite{cGAN,AuxOdena}.
Recent approaches \cite{BigGAN,VQ_VAE_2} have improved the image quality, resolution and diversity of generative models drastically.
However, there are two major drawbacks that limit their practical application: they are based on single-object datasets, and do not allow reconfiguration of individual aspects of the image to be generated.

\myparagraph{Layout-to-Image}
The direct layout-to-image task was first studied in Layout2Im \cite{Layout2Im} using a VAE \cite{VAE} based approach that could produce diverse 64$\times$64 pixel images by decomposing the representation of each object into a specified label and an unspecified (sampled) appearance vector.
LostGAN \cite{LostGAN,LostGANv2} allows better control over individual objects using a reconfigurable layout while keeping existing objects in the generated image unchanged.
This is achieved by providing individual latent style codes for each object, wherein one code for the whole image allows to generate diverse images from the same layout when the object codes are fixed.
We use LostGAN as our backbone to successfully address a fundamental problem: the inability to specify the appearance of individual objects using attributes.

\myparagraph{Scene-Graph-to-Image}
Scene graphs represent a scene of multiple objects using a graph structure where objects are encoded as nodes and edges represent the relationships between individual objects.
Due to their convenient and flexible structure, scene graphs have recently been employed in multiple image generation approaches \cite{Sg2Im,canonicalSg2Im,OCGAN,Grid2Im,PasteGAN}.
Typically, a graph convolution network (GCN) \cite{GCN} is used to predict a scene layout containing segmentation masks and bounding boxes for each object which is then used to generate an image.
However, scene graphs can be cumbersome to edit and do not allow to specify object locations directly on the image canvas.

\begin{figure*}[t]
\centering
\includegraphics[width=\linewidth]{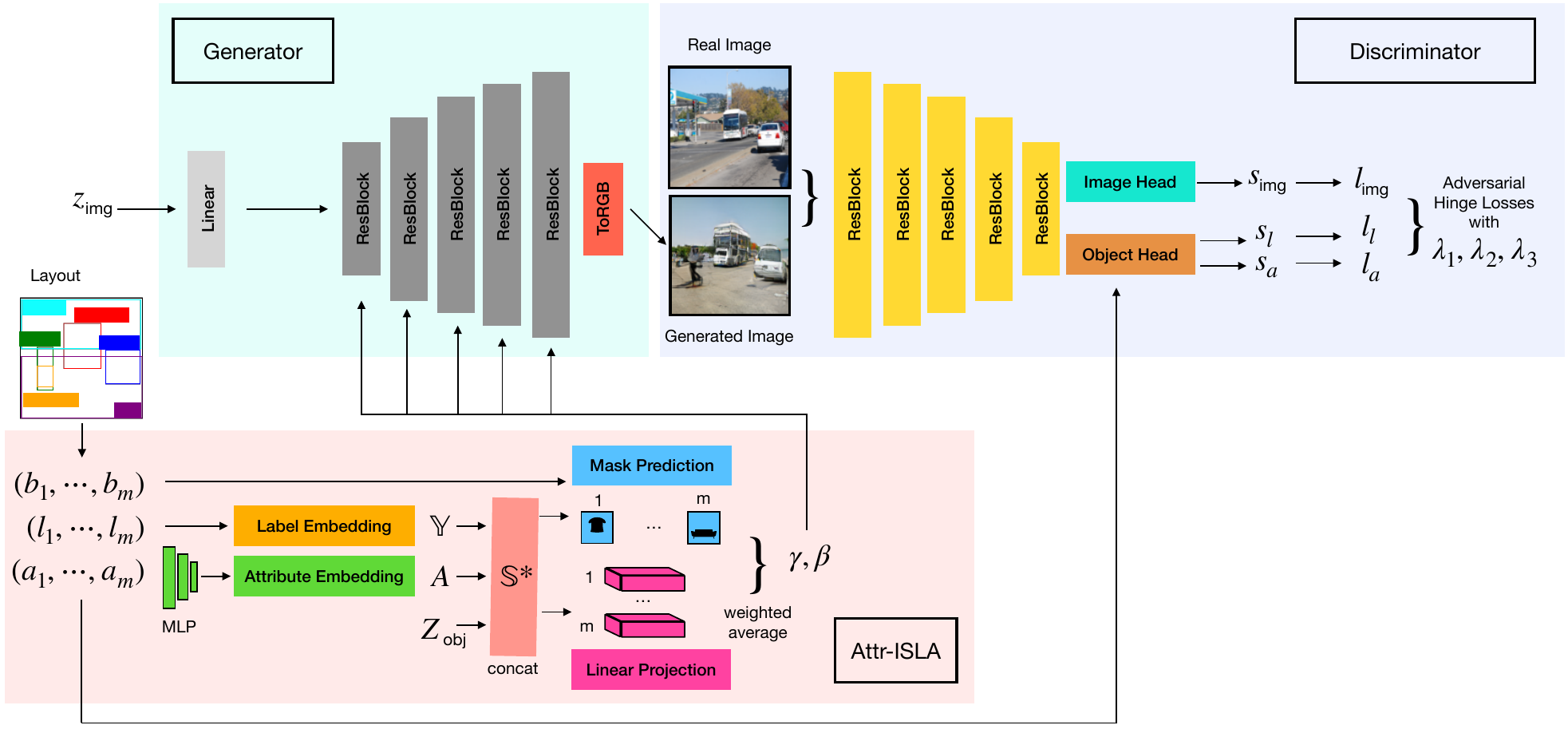}
   \caption{Illustration of our proposed method for attribute controlled image synthesis from reconfigurable layout and style.
   Given a layout of object positions, class labels and attributes, we compute affine transformation parameters $\gamma$ and $\beta$ to condition the generator.
   Using separate latent codes $Z_{\textrm{obj}}$ and $z_{\textrm{img}}$ for individual objects and image, respectively, enables our model to produce diverse images.
   The discriminator minimizes three adversarial hinge losses on image features, object-label features, and object-attribute features.}
\label{fig:model}
\end{figure*}

\myparagraph{Text-to-Image}
Textual descriptions provide an intuitive way for conditional image synthesis \cite{frolov2021adversarial}.
Current methods \cite{Reed2016,StackGAN,AttnGAN,DMGAN,OPGAN} first produce a text embedding which is then input to a multi-stage image generator.
In \cite{Hinz2019GeneratingMO,OPGAN}, additional layout information is used by adding an object pathway to learn the features of individual objects and control object locations.
Decomposing the task into predicting a semantic layout from text, and then generate images conditioned on both text and semantic layout has been explored in \cite{InferGAN,ObjGAN}.
Other works focus on disentangling content from style \cite{ControlGAN,Zhou2019}, and text-guided image manipulation \cite{TAGAN,ManiGAN}.
However, natural language can be ambiguous and textual descriptions are difficult to obtain.

\myparagraph{Usage of Attributes}
Early methods used attributes to generate outdoor scene \cite{karacan2016learning,StarGAN}, human face and bird images \cite{Attribute2Image}.
In contrast, our method can generate complex scene images containing multiple objects from a reconfigurable layout.
Most similar to our work are the methods proposed by Ke Ma \etal \cite{AttrLayout2Im} as an extension of \cite{AttrLayout2Im} using an auxiliary attribute classifier and explicit reconstruction loss for horizontally shifted objects, and \cite{pavllo2020controlling} which requires semantic instance masks.
To the best of our knowledge, \cite{AttrLayout2Im} is currently the only other direct layout-to-image method using attributes.
Our method improves upon \cite{AttrLayout2Im} in terms of visual quality, control and image resolution using a straightforward, yet effective approach built on \cite{LostGAN,LostGANv2}.

\section{Approach}
\label{sec:approach}

LostGAN \cite{LostGAN,LostGANv2} achieves remarkable results and control in the layout-to-image task, but it lacks the ability to specify the attributes of an object.
While an object class label defines the high-level category (\eg, ``car'', ``person'', ``dog'', ``building)'',  attributes refer to structural properties and appearance variations such as colors (\eg, ``blue'', ``yellow''), sentiment (\eg,  ``happy'',  ``angry''), and forms (\eg, ``round'', ``sliced'') which can be assigned to a variety of object classes \cite{VisualGenome}.
Although one could randomly sample many different object latent codes to generate diverse outputs, it does not allow to provide specific descriptions of the appearance to enable users to generate ``what they have in mind''.
To address this fundamental problem, we build upon \cite{LostGAN} and additionally condition individual objects on a set of attributes.
To that end, we create an attribute embedding, similar to the label embedding used in \cite{LostGAN}, and propose Attr-ISLA to compute affine transformation parameters which depend on object positions, class labels and attributes.
Furthermore, we utilize a separate attribute embedding to compute an additional adversarial hinge loss on object-attribute features.
See \autoref{fig:model} for an illustration of our method.

\subsection{Problem Formulation}
Given an image layout $L=\{(l_i, b_i, a_i)_{i=1}^{m}\}$ of $m$ objects, where each object is defined by a class label $l_i$, a bounding box $b_i$, and attributes $a_i$, the goal of our method is to generate an image $I$ with accurate positioned and recognizable objects which also correctly reflect their corresponding input attributes.
We use LostGAN \cite{LostGAN,LostGANv2} as our backbone, in which the overall style of the image is controlled by the latent $z_{\textrm{img}}$, and individual object styles are controlled by the set of latents $Z_\textrm{obj}=\{z_i\}_{i=1}^m$.
Latent codes are sampled from the standard normal distribution $\mathcal{N}(0,1)$.
Note, the instance object style codes $Z_\textrm{obj}$ are important even though attributes are provided to capture the challenging one-to-many mapping and enable the generation of diverse images (\eg, there are many possible images of a person wearing a blue shirt).
In summary, we want to find a generator function $G$ parameterized by $\Theta_G$ which captures the underlying conditional data distribution $p=(I | L, z_{\textrm{img}}, Z_\textrm{obj})$ such that we can use it to generate new, realistic samples.
Similar to \cite{LostGAN}, the task we are addressing in this work can hence be expressed more formally as in \autoref{eq:problem}
\begin{align}\label{eq:problem}
  I=G(L,z_{\textrm{img}}, Z_\textrm{obj}; \Theta_G),
\end{align}

where all components of the layout $L$ (i.e., class labels $l_i$, object positions $b_i$ and attributes $a_i$) are reconfigurable to allow fine-grained control of diverse images using the randomly sampled latents $z_{\textrm{img}}$ and $Z_\textrm{obj}$.
In other words, our goals are 1) to control the appearance of individual objects using attributes, but still be able to 2) reconfigure the layout and styles to generate diverse objects corresponding to the desired specification.

\subsection{Attribute ISLA (Attr-ISLA)}
Inspired by \cite{miyato2018cgans,BigGAN,StyleGAN}, the authors of \cite{LostGAN} extended the Adaptive Instance Normalization (AdaIN) \cite{StyleGAN} to object Instance-Sensitive and Layout-Aware feature Normalization (ISLA-Norm) to enable fine-grained and multi-object style control.
In order to gain control over the appearance of individual objects, we propose to additionally condition on object attributes using a simple, yet effective enhancement to the ISLA-Norm \cite{LostGAN}.
On a high-level, the channel-wise batch mean $\mu$ and variance $\sigma$ are computed as in BatchNorm \cite{BatchNorm} while the affine transformation parameters $\gamma$ and $\beta$ are instance-sensitive (class labels and attributes) and layout-aware (object positions) per sample.
Similar to \cite{LostGAN}, this is achieved in a multi-step process:

\emph{1) Label Embedding:}
Given one-hot encoded label vectors for $m$ objects with $d_l$ denoting the number of class labels, and $d_e$ the embedding dimension, the one-hot label matrix $Y$ of size $m \times d_l$ is transformed into the $m \times d_e$ label-to-vector matrix representation of labels $\mathbb{Y}=Y \cdot W$ using a learnable $d_l \times d_e$ size embedding matrix $W$.

\emph{2) Attribute Embedding:}
Given binary encoded attribute vectors for $m$ objects, an intermediate MLP is used to map the attributes into an $m \times d_e$ size attribute-to-vector matrix representation $A$.

\emph{3) Joint Label, Attribute \& Style Projection:}
The sampled object style noise matrix $Z_\textrm{obj}$ of size $m \times d_\textrm{noise}$ is concatenated with the label-to-vector matrix $\mathbb{Y}$ and attribute-to-vector matrix $A$ to obtain the $m \times (2 \cdot d_e + d_\textrm{noise})$ size embedding matrix $\mathbb{S}^* = (\mathbb{Y}, A, Z_\textrm{obj})$.
The embedding matrix $\mathbb{S}^*$, which now depends on the class labels, attributes and latent style codes, is used to compute object attribute-guided instance-sensitive channel-wise $\gamma$ and $\beta$ via linear projection using a learnable $(2 \cdot d_e + d_\textrm{noise}) \times 2C$ projection matrix, with $C$ denoting the number of channels.

\emph{4) Mask Prediction:}
A non-binary $s \times s$ mask is predicted for each object by a sub-network consisting of up-sample convolutions and a sigmoid transformation.
Next, the masks are resized to the corresponding bounding box sizes.

\emph{5) ISLA $\gamma$, $\beta$ Computation:}
The $\gamma$ and $\beta$ parameters are unsqueezed to their corresponding bounding boxes, weighted by the predicted masks, and finally added together with averaged sum used for overlapping regions.

Because the affine transformation parameters depend on individual objects in a sample (class labels, bounding boxes, attributes and styles), our AttrLostGAN achieves better and fine-grained control over the image generation process.
This allows the user to create an image iteratively and interactively by updating the layout, specifying attributes, and sampling latent codes.
We refer the reader to \cite{LostGAN} for more details on ISLA and \cite{LostGANv2} for an extended ISLA-Norm which integrates the learned masks at different stages in the generator.

\subsection{Architecture and Objective}
We use LostGAN \cite{LostGAN,LostGANv2} as our backbone without changing the general architecture of the ResNet \cite{ResNet} based generator and discriminator.
The discriminator $D(\cdot;\Theta_D)$ consists of three components: a shared ResNet backbone to extract features, an image head classifier, and an object head classifier.
Following the design of a separate label embedding to compute the object-label loss, we create a separate attribute embedding to compute the object-attribute loss to encourage the generator $G$ to produce objects with specified attributes.
Similar to \cite{LostGAN,LostGANv2}, the objective can be formulated as follows.
Given an image $I$, the discriminator predicts scores for the image ($s_\textrm{img}$), and average scores for the object-label ($s_{l}$) and object-attribute ($s_{a}$) features, respectively:
\begin{align} \label{eq:discriminator}
  (s_\textrm{img}, s_{l}, s_{a}) = D(I,L;\Theta_D)
\end{align}

We use the adversarial hinge losses
\begin{align}\label{eq:hinge}
  \mathcal{L}_{t}(I, L)=\left\{\begin{array}{ll}
\max \left(0,1-s_{t}\right) ; & \text { if } I \text { is real } \\
\max \left(0,1+s_{t}\right) ; & \text { if } I \text { is fake }
\end{array}\right.
\end{align}

where $t \in \{\textrm{img}, l, a\}$.
The objective can hence be written as
\begin{align}\label{eq:objective}
  \mathcal{L}(I, L)=\lambda_1 \mathcal{L}_\textrm{img}(I,L) + \lambda_2 \mathcal{L}_l(I,L) + \lambda_3 \mathcal{L}_a(I,L),
\end{align}

where $\lambda_1,\lambda_2,\lambda_3$ are trade-off parameters between image, object-label, and object-attribute quality.
The losses for the discriminator and generator can be written as
\begin{align}\label{eq:loss}
  \begin{aligned}
\mathcal{L}_D &= \mathbb{E}\left[\mathcal{L}\left(I^{\textrm{real}}, L\right)+\mathcal{L}\left(I^{\textrm{fake}}, L\right)\right] \\
\mathcal{L}_G &= -\mathbb{E}\left[\mathcal{L}\left(I^{\textrm{fake}}, L\right)\right]
\end{aligned}
\end{align}

We set $\lambda_1=0.1$, $\lambda_2=1.0$, and $\lambda_3=1.0$ to obtain our main results in \autoref{tab:vg}, and train our models for 200 epochs using a batch size of 128 on three NVIDIA V100 GPUs.
Both $\lambda_1$, and $\lambda_2$ are as in \cite{LostGAN}.
We use the Adam optimizer, with $\beta_1=0$, $\beta_2=0.999$, and learning rates $10^{-4}$ for both generator and discriminator.

\section{Experiments}
\label{sec:experiments}
Since we aim to gain fine-grained control of individual objects using attributes, we first create and experiment with a synthetic dataset to demonstrate the effectiveness of our approach before moving to the challenging Visual Genome \cite{VisualGenome} dataset.

\begin{figure*}[t]
\centering
\includegraphics[width=1.0\linewidth]{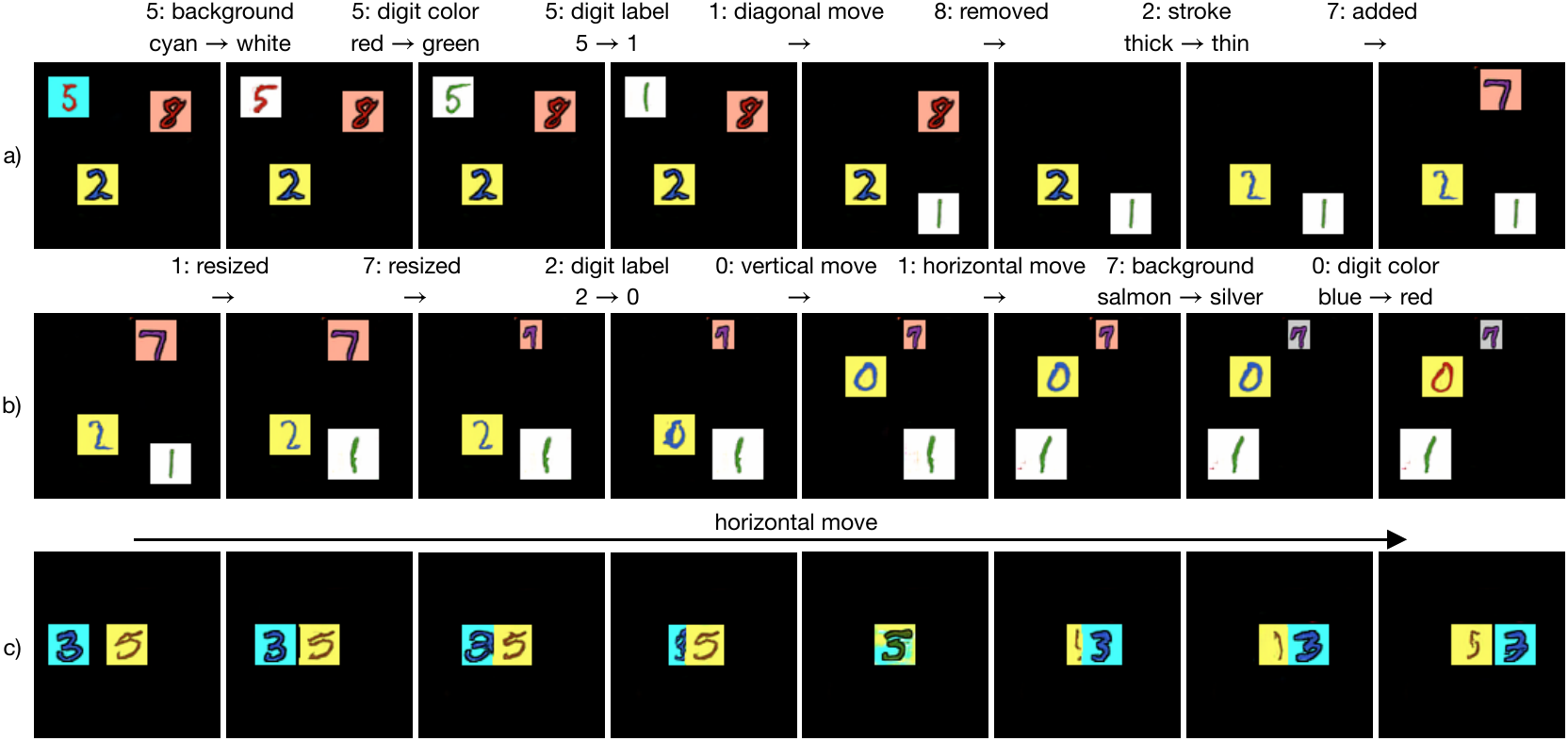}
   \caption{Iterative reconfiguration example on the MNIST Dialog based dataset (all images are generated by our model).
   In a) and b), we reconfigure various aspects of the (not shown) input layout to demonstrate controlled image generation.
   Our approach allows fine-grained control over individual objects with no or minimal changes to other parts of the image.
   During reconfiguration, we can sometimes observe small style changes, indicating partial entanglement of latent codes and specified attributes.
   In c), we horizontally shift one object showing that nearby and overlapping objects influence each other which might be a desirable feature to model interactions in more complex settings.
   }
\label{fig:dialogmnist}
\end{figure*}

\subsection{MNIST Dialog}

\myparagraph{Dataset}
We use the MNIST Dialog \cite{mnist_dialog} dataset and create an annotated layout-to-image dataset with attributes.
In MNIST Dialog each image is a 28$\times$28 pixel MNIST image with three additional attributes, i.e., digit color (red, blue, green, purple, or brown), background color (cyan, yellow, white, silver, or salmon), and style (thick or thin).
Starting from an empty 128$\times$128 image canvas, we randomly select, resize and place 3-8 images on it thereby creating an annotated layout-to-image with attributes dataset, where each ``object'' in the image is an image from MNIST Dialog.
While randomly placing the images on the canvas, we ensure that each image is sufficiently visible by allowing max.\ 40\% overlap between any two images.

\myparagraph{Results}
In \autoref{fig:dialogmnist}, we depict generated images using a corresponding layout.
Our model learned to generate sharp objects at the correct positions with corresponding labels and attributes, and we can successfully control individual object attributes without affecting other objects.
When reconfiguring one object we can sometimes observe slight changes in the style of how a digit is drawn, indicating that the variation provided by the object latent codes is not fully disentangled from the attribute specification.
However, we also observe that other objects remain unchanged, hence providing fine-grained control over the individual appearance.
We further show how two nearby or even overlapping objects can influence each other which might be necessary to model interacting objects in more complex settings.
We hypothesize this is due to the weighted average pool in ISLA which computes an average style for that position.

\begin{figure*}[t]
\centering
\includegraphics[width=1.0\linewidth]{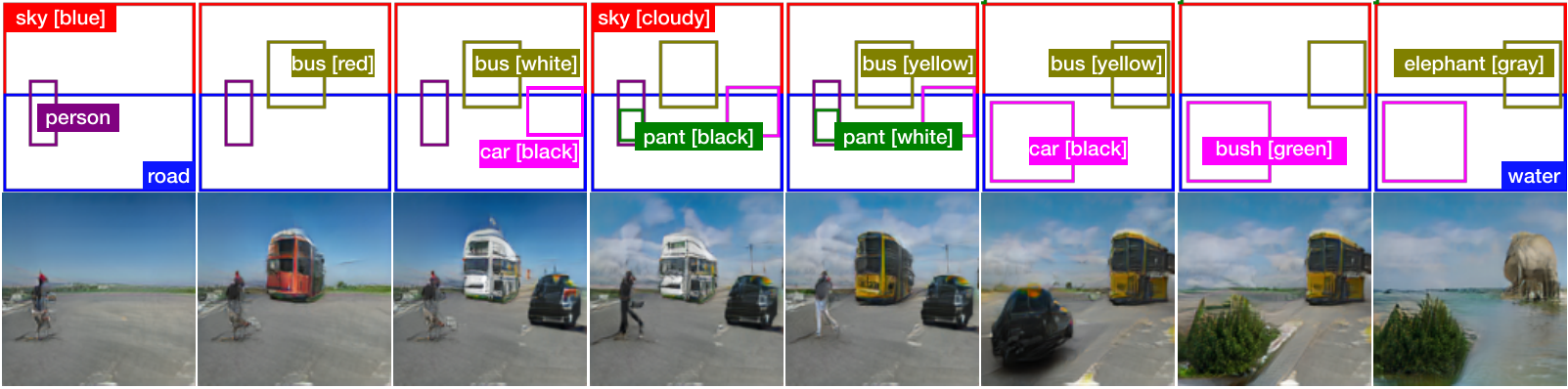}
   \caption{Our model can control the appearance of generated objects via attributes.
   \textit{From left to right:} add bus [red]; bus [red $\rightarrow$ white], add car [black]; sky [blue $\rightarrow$ cloudy], add pant [black]; bus [white $\rightarrow$ yellow], pant [black $\rightarrow$ white]; remove person, remove pant [white], reposition bus [yellow], reposition and resize car [black]; car [black] $\rightarrow$ bush [green]; road $\rightarrow$ water, bus [yellow] $\rightarrow$ elephant [gray].}
\label{fig:another_iterative}

\vspace{1em}
\includegraphics[width=1.0\linewidth]{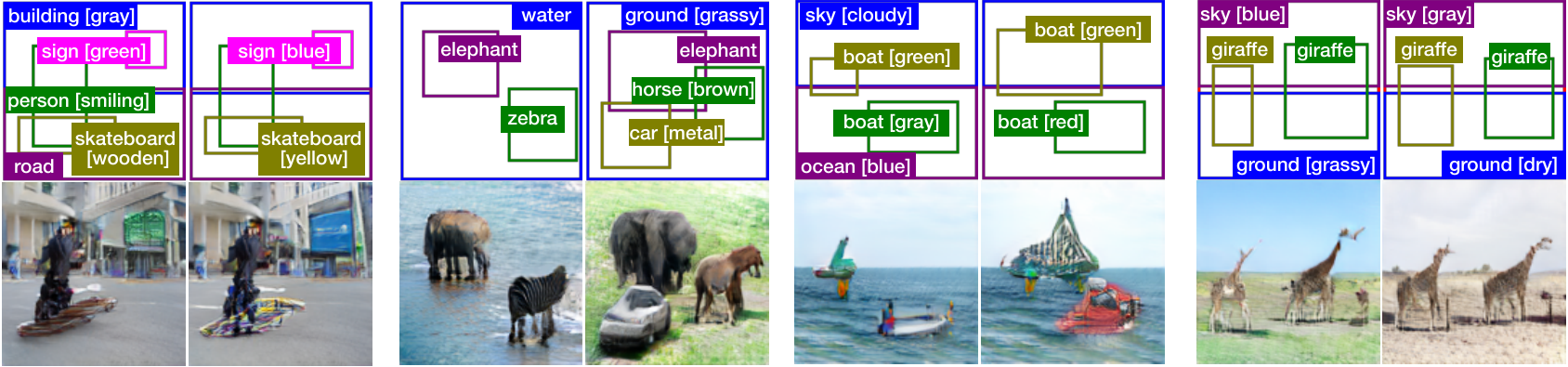}
   \caption{More reconfiguration examples.
   \textit{First pair:} sign [green $\rightarrow$ blue], skateboard [wooden $\rightarrow$ yellow].
   \textit{Second pair:} water $\rightarrow$ ground [grassy], zebra $\rightarrow$ horse [brown], repositioned horse [brown], resized elephant, added car [metal].
   \textit{Third pair:} boat [gray $\rightarrow$ red], resized boat [green].
   \textit{Fourth pair:} sky [blue $\rightarrow$ gray], ground [grassy $\rightarrow$ dry], resized both giraffes.}
\label{fig:reconfigs}

\vspace{1em}
\includegraphics[width=1.0\linewidth]{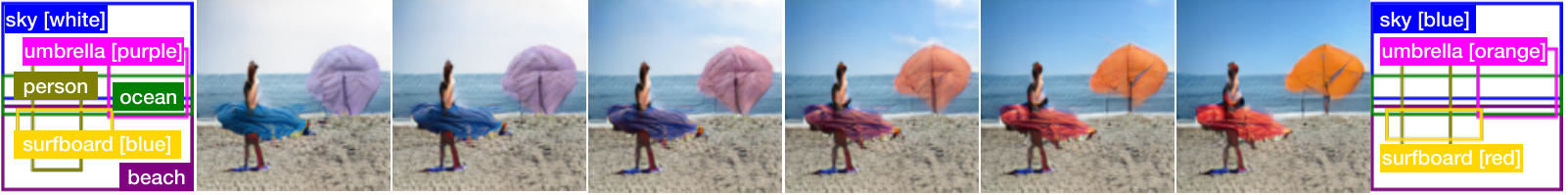}
   \caption{Generating images from a linear interpolation between attribute embeddings produces smooth transitions.
   From left to right, we interpolate between the following attribute specification: sky [white $\rightarrow$ blue], umbrella [purple $\rightarrow$ orange], surfboard [blue $\rightarrow$ red].
   }
\label{fig:linear_interpolation}
\end{figure*}

\subsection{Visual Genome}

\myparagraph{Dataset}
Finally, we apply our proposed method to the challenging Visual Genome \cite{VisualGenome} dataset.
Following the setting in \cite{Sg2Im,LostGAN}, we pre-process and split the dataset by removing small and infrequent objects, resulting in 62,565 training, 5,062 validation, and 5,096 testing images, with 3 to 30 objects from 178 class labels per image.
We filter all available attributes to include only such that appear at least 2,000 times, and allow up to 30 attributes per image.

\myparagraph{Metrics}
Evaluating generative models is challenging because there are many aspects that would resemble a good model such as visual realism, diversity, and sharp objects \cite{Theis2015ANO,Borji2018ProsAC}.
Additionally, a good layout-to-image model should generate objects of the specified class labels and attributes at their corresponding locations.
Hence, we choose multiple metrics to evaluate our model and compare with baselines.

To evaluate the image quality and diversity, we use the IS \cite{IS} and FID \cite{FID}.
To assess the visual quality of individual objects we choose the SceneFID \cite{OCGAN} which corresponds to the FID applied on cropped objects as defined by the bounding boxes.
Similarly, we propose to apply the IS on generated object crops, denoted as SceneIS.
As in \cite{LostGAN}, we use the CAS \cite{CAS}, which measures how well an object classifier trained on generated can perform on real image crops.
Note, this is different to the classification accuracy as used in \cite{Layout2Im,AttrLayout2Im}, which is trained on real and tested on generated data, and hence might overlook the diversity of generated images \cite{LostGANv2}.
Additionally, and in the same spirit as CAS, we report the micro F1 (Attr-F1) by training multi-label classification networks to evaluate the attribute quality by training on generated and test on real object crops.
As in \cite{Layout2Im,LostGAN,LostGANv2,AttrLayout2Im}, we adopt the LPIPS metric as the Diversity Score (DS) \cite{DS} to compute the perceptual similarity between two sets of images generated from the same layout in the testing set.

\begin{table*}
\small
\begin{center}
    \begin{tabular}{r c c c c c c c }
    \toprule
    Method & IS $\uparrow$ & SceneIS $\uparrow$ & FID $\downarrow$ & SceneFID $\downarrow$ & DS ($\uparrow$) & CAS $\uparrow$ & Attr-F1 $\uparrow$  \\
    \midrule
    Real Images (128$\times$128)                 & 23.50 $\pm$ 0.71 & 13.43 $\pm$ 0.33  & 11.93    & 2.46  & -                & 46.22 & 15.77  \\
    \midrule
    Layout2Im \cite{Layout2Im} (64$\times$64)      & 8.10 $\pm$ 0.10  & -                 & 40.07    & -     & 0.17 $\pm$ 0.09  & -     & -       \\
    LostGANv1 \cite{LostGAN} (128$\times$128)           & 10.30 $\pm$ 0.19 & 9.07 $\pm$ 0.12 & 35.20    & 11.06 & 0.47 $\pm$ 0.09  & 31.04 & 11.25    \\
    LostGANv2 \cite{LostGANv2} (128$\times$128) & 10.25 $\pm$ 0.20 & 9.15 $\pm$ 0.22 & 34.77 & 15.25 & 0.42 $\pm$ 0.09 & 30.97 & 11.38 \\
    Ke Ma \etal \cite{AttrLayout2Im} (128$\times$128) & 9.57 $\pm$ 0.18 & 8.17 $\pm$ 0.13 & 43.26 & 16.16 & 0.30 $\pm$ 0.11  & \bftab 33.09 & 12.62    \\
    \textit{AttrLostGANv1} (128$\times$128)               & 10.68 $\pm$ 0.43 & 9.24 $\pm$ 0.12 & 32.93    & 8.71  & 0.40 $\pm$ 0.11  & 32.11 & 13.64     \\
    \textit{AttrLostGANv2} (128$\times$128)             & \bftab 10.81 $\pm$ 0.22 & \bftab 9.46 $\pm$ 0.13 & \bftab 31.57    & \bftab 7.78  & 0.28 $\pm$ 0.10  & 32.90 & \bftab 14.61     \\
    \midrule
    \midrule
    Real Images (256$\times$256)                    & 31.41 $\pm$ 1.15 & 19.58 $\pm$ 0.27 & 12.41 & 2.78 & - & 50.94 & 17.80 \\
    \midrule
    LostGANv2 \cite{LostGANv2} (256$\times$256) & \bftab 14.88 $\pm$ 0.25 & 11.87 $\pm$ 0.16 & \bftab 35.03 & 18.87 & 0.53 $\pm$ 0.10 & \bftab 35.80 & 11.97 \\
    \textit{AttrLostGANv2} (256$\times$256)            &  14.25 $\pm$ 0.61 & \bftab 11.96 $\pm$ 0.36 & 35.73 & \bftab 14.76 &  0.45 $\pm$ 0.11 & 35.36 & \bftab 14.49  \\
    \bottomrule
    \end{tabular}
    \end{center}
    \caption{Results on Visual Genome. Our models (in italics) achieve the best scores on most metrics, and AttrLostGANv2 is considerably better than AttrLostGANv1.
    When trained on higher resolution, our model performs better in terms of object and attribute quality, while achieving similar scores on image quality metrics.
    Note, a lower diversity (DS) is expected due to the specified attributes.
    }
\label{tab:vg}
\end{table*}

\begin{figure*}[ht]
\centering
\includegraphics[width=1.0\linewidth]{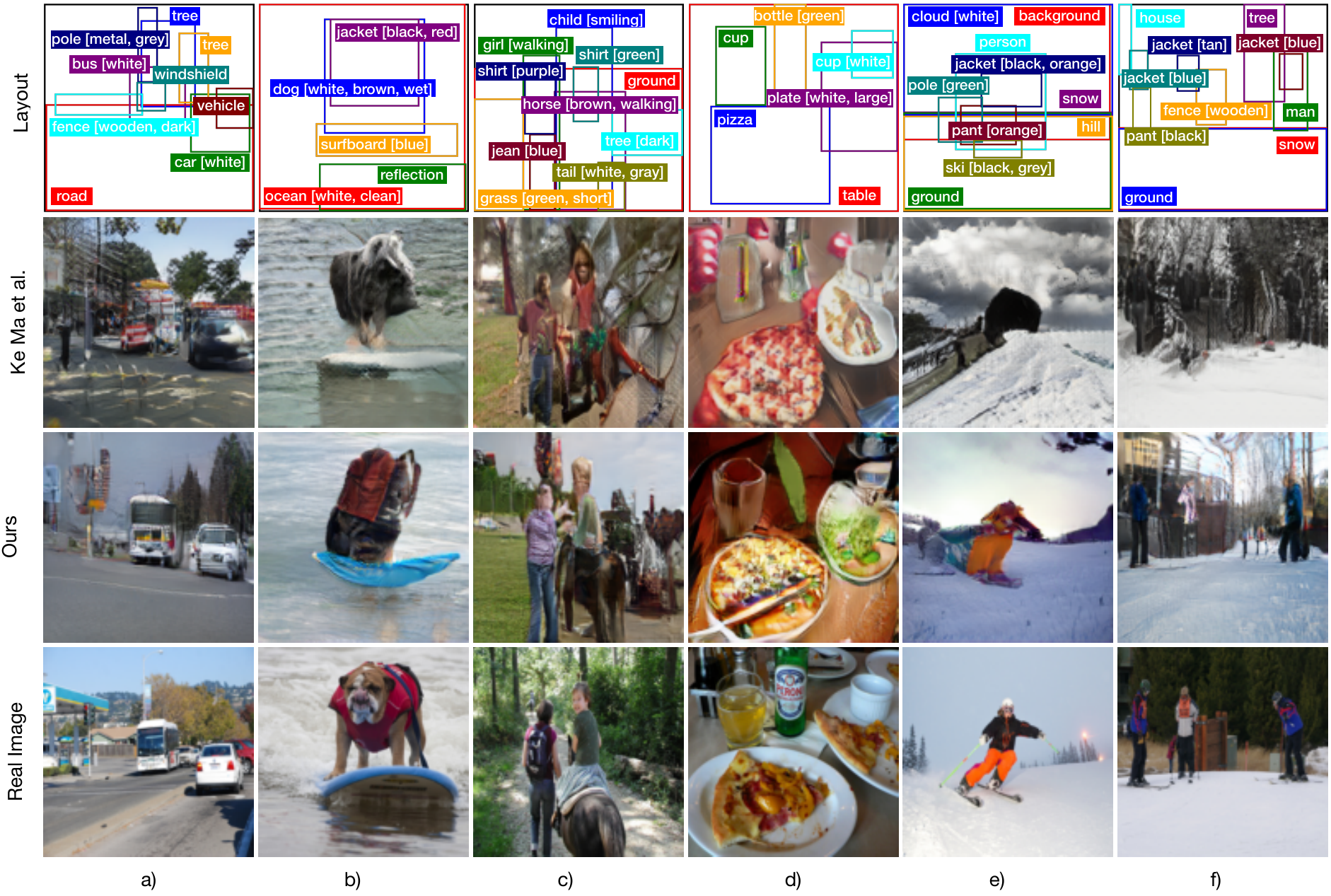}
   \caption{Visual comparison between images generated by our AttrLostGANv1 and Ke Ma \etal \cite{AttrLayout2Im} using the layouts shown in the first row.
   Our images are consistently better at reflecting the input attributes and individual objects have more details and better texture.
   For example, a) bus [white], b) jacket [black, red], c) shirt [purple], d) plate [white, large], e) pant [orange], f) jacket [blue].}
\label{fig:visualgenome}
\end{figure*}

\myparagraph{Qualitative Results}
\autoref{fig:iterative} and \autoref{fig:another_iterative} depict examples of attribute controlled image generation from reconfigurable layout.
Our model provides a novel way to iteratively reconfigure the properties of individual objects to generate images of complex scenes without affecting other parts of the image.
\autoref{fig:reconfigs} shows more examples in which we reconfigure individual objects by changing attributes, class labels, object position and size.
In \autoref{fig:linear_interpolation}, we linearly interpolate between two sets of attributes for the same layout.
Our model learns a smooth transition between attributes.
In \autoref{fig:visualgenome} we compare generated images between our AttrLostGANv1 and \cite{AttrLayout2Im} using layouts from the testing set.
As can be seen, generating realistic images of complex scenes with multiple objects is still very difficult.
Although the images generated by our model look more realistic, individual objects and details such as human faces are hard to recognize.
In terms of attribute control, our images better depict the input specifications in general.

\myparagraph{Quantitative Results}
\autoref{tab:vg} shows quantitative results.
We train two variants of our approach: AttrLostGANv1 which is based on \cite{LostGAN}, and AttrLostGANv2 which is based on \cite{LostGANv2}.
We compare against the recent and only other direct layout-to-image baseline proposed by Ke Ma \etal \cite{AttrLayout2Im}, which is an extension of Layout2Im \cite{Layout2Im} that can be conditioned on optional attributes.
Since no pre-trained model was available at the official codebase of \cite{AttrLayout2Im}, we used the open-sourced code to train a model.
For fair comparison, we evaluate all models trained by us.
Our models achieve the best scores across most metrics, and AttrLostGANv2 is considerably better than AttrLostGANv1.
\cite{AttrLayout2Im} reaches a competing performance on attribute control, but inferior image and object quality.
For example, our method increases the SceneIS from 8.17 to 9.46, and lowers the FID from 43.26 to 31.57.
Furthermore, our method is better at generating the appearance specified by the attributes as indicated by the improvement of Attr-F1 from 12.62 to 14.61.
In terms of CAS, our model performs slightly worse than \cite{AttrLayout2Im}, which might be due to the explicit attribute classifier used in \cite{AttrLayout2Im} during training.
Building upon \cite{LostGANv2} our method can also generate higher resolution images (256$\times$256 compared to 128$\times$128).
By specifying attributes, a decreased DS is expected but we include it for completeness.

\begin{table}
\begin{center}
    \resizebox{%
      \ifdim\width>\columnwidth
        \columnwidth
      \else
        \width
      \fi}{!}{%
    \begin{tabular}{l c c c c }
    \toprule
    Method          & FID $\downarrow$ & SceneFID $\downarrow$ & CAS $\uparrow$ & Attr-F1 $\uparrow$  \\
    \midrule
    Real Images                & 11.93          & 2.46             & 46.22           & 15.77             \\
    \midrule
    LostGAN \cite{LostGAN}     & 35.20          & 11.06            & 31.04           & 11.25             \\
    \midrule
    + attr. context            & \bftab 30.74   & 9.74             & 32.94           & 12.97             \\
    \midrule
    + loss $\lambda_3=0.1$     & 34.16          & 9.95             & \bftab 33.78  & 12.85             \\
    + loss $\lambda_3=0.5$     & 33.66          & 9.24             & 32.47           & 13.52             \\
    + loss $\lambda_3=1.0$     & 31.82          & 9.54             & 31.53           & 13.56             \\
    + loss $\lambda_3=2.0$     & 34.84          & \bftab 8.55      & 29.29           & 13.56             \\
    \midrule
    + MLP depth=1              & 35.42          & 8.87             & 32.30           & 12.81             \\
    + MLP depth=2              & 32.96          & 9.90             & 32.05           & 13.38             \\
    + MLP depth=3 (ours)       & 32.93          & \underline{8.71} & 32.11           & \underline{13.64} \\
    + MLP depth=4              & 36.76          & 10.21            & 31.71           & \bftab 14.88    \\
    \bottomrule
    \end{tabular}}
    \end{center}
    \caption{
    Results of our ablation study on AttrLostGANv1. We ablate the weight $\lambda_3$ used in the adversarial hinge loss on object-attribute features, with higher values of $\lambda_3$ leading to higher Attr-F1 and lower values decreasing the image quality.
    The ablation is also performed on the intermediate depth of the attribute MLP (setting $\lambda_3=1.0$).
    A medium depth MLP with three hidden layers achieves the best overall performance.
    }
\label{tab:ablation}
\end{table}

\myparagraph{Ablation Study}
We also perform ablations of our main changes to the LostGAN \cite{LostGAN} backbone, see \autoref{tab:ablation}.
Starting from LostGAN we add attribute information to the generator and already gain an improvement over the baseline in terms of image quality, object discriminability, as well as attribute information.
We ablate the additional adversarial hinge loss on object-attribute features $\lambda_3$.
A higher $\lambda_3$ leads to better Attr-F1, but decreased image and object quality.
Interestingly, a high $\lambda_3=2.0$ achieves the best SceneFID on object crops, while the image quality in terms of FID is worst.
Although we only have to balance three weights, our results show that there exists a trade-off between image and object quality.
We choose $\lambda_3=1.0$ for all remaining experiments and ablate the depth of the intermediate attribute MLP which is used to compute the attribute-to-vector matrix representation for all objects.
While a shallow MLP leads to a decreased performance, a medium deep MLP with three hidden layers achieves the best overall performance.

\subsection{Discussion}
Our approach takes an effective step towards reconfigurable, and controlled image generation from layout of complex scenes.
Our model provides unprecedented control over the appearance of individual objects without affecting the overall image.
Although the quantitative as well as visual results are promising current approaches require attribute annotations which are time-consuming to obtain.
While the attribute control is strong when fixing the object locations, as demonstrated in our results, the object styles can change when target objects are nearby or overlap.
We hypothesize that this might be due to the average pool in ISLA when combining label and attribute features of individual objects and might hence lead to entangled representations.
At the same time, such influence might be desirable to model object interactions in complex settings.
Despite clearly improving upon previous methods both quantitatively and qualitatively, current models are still far from generating high-resolution, realistic images of complex scenes with multiple interacting objects which limits their practical application.

\section{Conclusion}
\label{sec:conclusion}
In this paper, we proposed AttrLostGAN, an approach for attribute controlled image generation from reconfigurable layout and style.
Our method successfully addresses a fundamental problem by allowing users to intuitively change the appearance of individual object details without changing the overall image or affecting other objects.
We created and experimented on a synthetic dataset based on MNIST Dialog to analyze and demonstrate the effectiveness of our approach.
Further, we evaluated our method against the recent, and only other baseline on the challenging Visual Genome dataset both qualitatively and quantitatively.
We find that our approach not only outperforms the existing method in most common measures while generating higher resolution images, but also that it provides users with intuitive control to update the generated image to their needs.
In terms of future work, our first steps are directed towards enhancing the image quality and resolution.
We would also like to investigate unsupervised methods to address the need of attribute annotations and whether we can turn attribute labels into textual descriptions.

\myparagraph{Acknowledgments}
This work was supported by the BMBF projects
ExplAINN (Grant 01IS19074),
XAINES (Grant 01IW20005),
the NVIDIA AI Lab (NVAIL)
and the TU Kaiserslautern PhD program.

\clearpage

{\small
\bibliographystyle{ieee_fullname}
\bibliography{egbib}

\begin{thebibliography}{10}\itemsep=-1pt

\bibitem{Grid2Im}
Oron Ashual and Lior Wolf.
\newblock Specifying object attributes and relations in interactive scene
  generation.
\newblock In {\em Proceedings of the IEEE International Conference on Computer
  Vision}, pages 4561--4569, 2019.

\bibitem{Borji2018ProsAC}
Ali Borji.
\newblock Pros and cons of gan evaluation measures.
\newblock {\em Computer Vision and Image Understanding}, 179:41--65, 2018.

\bibitem{BigGAN}
Andrew Brock, Jeff Donahue, and Karen Simonyan.
\newblock Large scale gan training for high fidelity natural image synthesis.
\newblock In {\em International Conference on Learning Representations}, 2018.

\bibitem{casanova2020generating}
Arantxa Casanova, Michal Drozdzal, and Adriana Romero-Soriano.
\newblock Generating unseen complex scenes: are we there yet?
\newblock {\em arXiv:2012.04027}, 2020.

\bibitem{Choi2020FromIT}
Hyeong-Seok Choi, Chang-Dae Park, and Kyogu Lee.
\newblock From inference to generation: End-to-end fully self-supervised
  generation of human face from speech.
\newblock In {\em International Conference on Learning Representations}, 2020.

\bibitem{StarGAN}
Yunjey Choi, Minje Choi, Munyoung Kim, Jung-Woo Ha, Sunghun Kim, and Jaegul
  Choo.
\newblock Stargan: Unified generative adversarial networks for multi-domain
  image-to-image translation.
\newblock In {\em Proceedings of the IEEE Computer Vision and Pattern
  Recognition}, pages 8789--8797, 2018.

\bibitem{Dong2017}
Hao Dong, Simiao Yu, Chao Wu, and Yike Guo.
\newblock Semantic image synthesis via adversarial learning.
\newblock In {\em Proceedings of the IEEE International Conference on Computer
  Vision}, 2017.

\bibitem{frolov2021adversarial}
Stanislav Frolov, Tobias Hinz, Federico Raue, J{\"o}rn Hees, and Andreas
  Dengel.
\newblock Adversarial text-to-image synthesis: A review.
\newblock {\em arXiv:2101.09983}, 2021.

\bibitem{GAN}
Ian~J. Goodfellow, Jean Pouget-Abadie, Mehdi Mirza, Bing Xu, David
  Warde-Farley, Sherjil Ozair, Aaron~C. Courville, and Yoshua Bengio.
\newblock Generative adversarial nets.
\newblock In {\em Advances in Neural Information Processing Systems}, pages
  2672--2680, 2014.

\bibitem{ResNet}
Kaiming He, Xiangyu Zhang, Shaoqing Ren, and Jian Sun.
\newblock Deep residual learning for image recognition.
\newblock {\em Proceedings of the IEEE Computer Vision and Pattern
  Recognition}, pages 770--778, 2016.

\bibitem{GCN}
Mikael Henaff, Joan Bruna, and Yann LeCun.
\newblock Deep convolutional networks on graph-structured data.
\newblock {\em arXiv:1506.05163}, 2015.

\bibitem{canonicalSg2Im}
Roei Herzig, Amir Bar, Huijuan Xu, Gal Chechik, Trevor Darrell, and Amir
  Globerson.
\newblock Learning canonical representations for scene graph to image
  generation.
\newblock In {\em European Conference on Computer Vision}, 2020.

\bibitem{FID}
Martin Heusel, Hubert Ramsauer, Thomas Unterthiner, Bernhard Nessler, and Sepp
  Hochreiter.
\newblock Gans trained by a two time-scale update rule converge to a local nash
  equilibrium.
\newblock In {\em Advances in Neural Information Processing Systems}, pages
  6626--6637, 2017.

\bibitem{Hinz2019GeneratingMO}
Tobias Hinz, Stefan Heinrich, and Stefan Wermter.
\newblock Generating multiple objects at spatially distinct locations.
\newblock In {\em International Conference on Learning Representations}, 2019.

\bibitem{OPGAN}
Tobias Hinz, Stefan Heinrich, and Stefan Wermter.
\newblock Semantic object accuracy for generative text-to-image synthesis.
\newblock {\em IEEE Transactions on Pattern Analysis and Machine Intelligence},
  2020.

\bibitem{InferGAN}
Seunghoon Hong, Dingdong Yang, Jongwook Choi, and Honglak Lee.
\newblock Inferring semantic layout for hierarchical text-to-image synthesis.
\newblock In {\em Proceedings of the IEEE Computer Vision and Pattern
  Recognition}, pages 7986--7994, 2018.

\bibitem{BatchNorm}
Sergey Ioffe and Christian Szegedy.
\newblock Batch normalization: Accelerating deep network training by reducing
  internal covariate shift.
\newblock In {\em International Conference on Machine Learning}, pages
  448--456, 2015.

\bibitem{Isola_2017_CVPR}
Phillip Isola, Jun-Yan Zhu, Tinghui Zhou, and Alexei~A. Efros.
\newblock Image-to-image translation with conditional adversarial networks.
\newblock In {\em Proceedings of the IEEE Computer Vision and Pattern
  Recognition}, pages 1125--1134, 2016.

\bibitem{Sg2Im}
Justin Johnson, Agrim Gupta, and Li Fei-Fei.
\newblock Image generation from scene graphs.
\newblock In {\em Proceedings of the IEEE Computer Vision and Pattern
  Recognition}, pages 1219--1228, 2018.

\bibitem{karacan2016learning}
Levent Karacan, Zeynep Akata, Aykut Erdem, and Erkut Erdem.
\newblock Learning to generate images of outdoor scenes from attributes and
  semantic layouts.
\newblock {\em arXiv:1612.00215}, 2016.

\bibitem{StyleGAN}
Tero Karras, Samuli Laine, and Timo Aila.
\newblock A style-based generator architecture for generative adversarial
  networks.
\newblock In {\em Proceedings of the IEEE Computer Vision and Pattern
  Recognition}, pages 4401--4410, 2018.

\bibitem{VAE}
Diederik~P. Kingma and Max Welling.
\newblock Auto-encoding variational bayes.
\newblock {\em CoRR}, arXiv:1312.6114, 2013.

\bibitem{VisualGenome}
Ranjay Krishna, Yuke Zhu, Oliver Groth, Justin Johnson, Kenji Hata, Joshua
  Kravitz, Stephanie Chen, Yannis Kalantidis, Li-Jia Li, David~A Shamma, et~al.
\newblock Visual genome: Connecting language and vision using crowdsourced
  dense image annotations.
\newblock {\em International Journal of Computer Vision}, 123(1):32--73, 2017.

\bibitem{ManiGAN}
Bowen Li, Xiaojuan Qi, Thomas Lukasiewicz, and Philip~HS Torr.
\newblock Manigan: Text-guided image manipulation.
\newblock In {\em Proceedings of the IEEE Computer Vision and Pattern
  Recognition}, pages 7880--7889, 2020.

\bibitem{ControlGAN}
Bowen Li, Xiaojuan Qi, Thomas Lukasiewicz, and Philip H.~S. Torr.
\newblock Controllable text-to-image generation.
\newblock {\em Advances in Neural Information Processing Systems}, 2019.

\bibitem{ObjGAN}
Wenbo Li, Pengchuan Zhang, Lei Zhang, Qiuyuan Huang, Xiaodong He, Siwei Lyu,
  and Jianfeng Gao.
\newblock Object-driven text-to-image synthesis via adversarial training.
\newblock In {\em Proceedings of the IEEE Computer Vision and Pattern
  Recognition}, pages 12166--12174, 2019.

\bibitem{AttrLayout2Im}
Ke Ma, Bo Zhao, and Leonid Sigal.
\newblock Attribute-guided image generation from layout.
\newblock In {\em British Machine Vision Virtual Conference}, 2020.
\newblock arXiv:2008.11932.

\bibitem{cGAN}
Mehdi Mirza and Simon Osindero.
\newblock Conditional generative adversarial nets.
\newblock {\em arXiv:1411.1784}, 2014.

\bibitem{miyato2018cgans}
Takeru Miyato and Masanori Koyama.
\newblock cgans with projection discriminator.
\newblock {\em arXiv:1802.05637}, 2018.

\bibitem{TAGAN}
Seonghyeon Nam, Yunji Kim, and Seon~Joo Kim.
\newblock Text-adaptive generative adversarial networks: Manipulating images
  with natural language.
\newblock In {\em Advances in Neural Information Processing Systems}, page
  42–51, 2018.

\bibitem{AuxOdena}
Augustus Odena, Christopher Olah, and Jonathon Shlens.
\newblock Conditional image synthesis with auxiliary classifier gans.
\newblock In {\em International Conference on Machine Learning}, pages
  2642--2651, 2016.

\bibitem{pavllo2020controlling}
Dario Pavllo, Aurelien Lucchi, and Thomas Hofmann.
\newblock Controlling style and semantics in weakly-supervised image
  generation.
\newblock In {\em European Conference on Computer Vision}, pages 482--499,
  2020.

\bibitem{CAS}
Suman Ravuri and Oriol Vinyals.
\newblock Classification accuracy score for conditional generative models.
\newblock In {\em Advances in Neural Information Processing Systems}, pages
  12268--12279, 2019.

\bibitem{VQ_VAE_2}
Ali Razavi, Aaron van~den Oord, and Oriol Vinyals.
\newblock Generating diverse high-fidelity images with vq-vae-2.
\newblock In {\em Advances in Neural Information Processing Systems}, pages
  14866--14876, 2019.

\bibitem{Reed2016}
Scott~E. Reed, Zeynep Akata, Xinchen Yan, Lajanugen Logeswaran, Bernt Schiele,
  and Honglak Lee.
\newblock Generative adversarial text to image synthesis.
\newblock In {\em International Conference on Machine Learning}, pages
  1060--1069, 2016.

\bibitem{IS}
Tim Salimans, Ian Goodfellow, Wojciech Zaremba, Vicki Cheung, Alec Radford, and
  Xi Chen.
\newblock Improved techniques for training gans.
\newblock In {\em Advances in Neural Information Processing Systems}, pages
  2234--2242, 2016.

\bibitem{mnist_dialog}
Paul~Hongsuck Seo, Andreas Lehrmann, Bohyung Han, and Leonid Sigal.
\newblock Visual reference resolution using attention memory for visual dialog.
\newblock In {\em Advances in neural information processing systems}, pages
  3719--3729, 2017.

\bibitem{LostGAN}
Wei Sun and Tianfu Wu.
\newblock Image synthesis from reconfigurable layout and style.
\newblock In {\em Proceedings of the IEEE International Conference on Computer
  Vision}, 2019.

\bibitem{LostGANv2}
Wei Sun and Tianfu Wu.
\newblock Learning layout and style reconfigurable gans for controllable image
  synthesis.
\newblock {\em arXiv:2003.11571}, 2020.

\bibitem{OCGAN}
Tristan Sylvain, Pengchuan Zhang, Yoshua Bengio, R~Devon Hjelm, and Shikhar
  Sharma.
\newblock Object-centric image generation from layouts.
\newblock {\em arXiv:2003.07449}, 2020.

\bibitem{Theis2015ANO}
Lucas Theis, A{\"a}ron van~den Oord, and Matthias Bethge.
\newblock A note on the evaluation of generative models.
\newblock {\em CoRR}, arXiv:1511.01844, 2015.

\bibitem{Wang2017HighResolutionIS}
Ting-Chun Wang, Ming-Yu Liu, Jun-Yan Zhu, Andrew Tao, Jan Kautz, and Bryan
  Catanzaro.
\newblock High-resolution image synthesis and semantic manipulation with
  conditional gans.
\newblock {\em Proceedings of the IEEE Computer Vision and Pattern
  Recognition}, pages 8798--8807, 2017.

\bibitem{wang2020domain}
Xinsheng Wang, Tingting Qiao, Jihua Zhu, Alan Hanjalic, and O. Scharenborg.
\newblock S2igan: Speech-to-image generation via adversarial learning.
\newblock In {\em INTERSPEECH}, 2020.

\bibitem{AttnGAN}
Tao Xu, Pengchuan Zhang, Qiuyuan Huang, Han Zhang, Zhe Gan, Xiaolei Huang, and
  Xiaodong He.
\newblock Attngan: Fine-grained text to image generation with attentional
  generative adversarial networks.
\newblock In {\em Proceedings of the IEEE Computer Vision and Pattern
  Recognition}, pages 1316--1324, 2017.

\bibitem{Attribute2Image}
Xinchen Yan, Jimei Yang, Kihyuk Sohn, and Honglak Lee.
\newblock Attribute2image: Conditional image generation from visual attributes.
\newblock In {\em European Conference on Computer Vision}, pages 776--791,
  2016.

\bibitem{PasteGAN}
LI Yikang, Tao Ma, Yeqi Bai, Nan Duan, Sining Wei, and Xiaogang Wang.
\newblock Pastegan: A semi-parametric method to generate image from scene
  graph.
\newblock In {\em Advances in Neural Information Processing Systems}, pages
  3948--3958, 2019.

\bibitem{StackGAN}
Han Zhang, Tao Xu, Hongsheng Li, Shaoting Zhang, Xiaogang Wang, Xiaolei Huang,
  and Dimitris~N. Metaxas.
\newblock Stackgan++: Realistic image synthesis with stacked generative
  adversarial networks.
\newblock {\em IEEE Transactions on Pattern Analysis and Machine Intelligence},
  41:1947--1962, 2017.

\bibitem{DS}
Richard Zhang, Phillip Isola, Alexei~A. Efros, Eli Shechtman, and Oliver Wang.
\newblock The unreasonable effectiveness of deep features as a perceptual
  metric.
\newblock In {\em Proceedings of the IEEE Computer Vision and Pattern
  Recognition}, 2018.

\bibitem{Layout2Im}
Bo Zhao, Lili Meng, Weidong Yin, and Leonid Sigal.
\newblock Image generation from layout.
\newblock In {\em Proceedings of the IEEE Computer Vision and Pattern
  Recognition}, 2019.

\bibitem{Zhou2019}
Xingran Zhou, Siyu Huang, Bin Li, Yingming Li, Jiachen Li, and Zhongfei Zhang.
\newblock Text guided person image synthesis.
\newblock In {\em Proceedings of the IEEE Computer Vision and Pattern
  Recognition}, pages 3663--3672, 2019.

\bibitem{Zhu_2017_ICCV}
Jun-Yan Zhu, Taesung Park, Phillip Isola, and Alexei~A. Efros.
\newblock Unpaired image-to-image translation using cycle-consistent
  adversarial networks.
\newblock In {\em Proceedings of the IEEE International Conference on Computer
  Vision}, pages 2223--2232, 2017.

\bibitem{DMGAN}
Minfeng Zhu, Pingbo Pan, Wei Chen, and Yi Yandg.
\newblock Dm-gan: Dynamic memory generative adversarial networks for
  text-to-image synthesis.
\newblock In {\em Proceedings of the IEEE Computer Vision and Pattern
  Recognition}, pages 5802--5810, 2019.

\end{thebibliography}
}

\clearpage

\appendix

\section{Training Details}
All our models were trained for 200 epochs using a batch size of 128 on three NVIDIA V100 GPUs.
Training our 128$\times$128 models took roughly four days, while our 256$\times$256 model took roughly 10 days.
We set $\lambda_1=0.1$, $\lambda_2=1.0$, and $\lambda_3=1.0$ to obtain our main results.
Both $\lambda_1$, and $\lambda_2$ are as in \cite{LostGAN}.
We use the Adam optimizer, with $\beta_1=0$, $\beta_2=0.999$, and learning rates $10^{-4}$ for both generator and discriminator.
To obtain the results for the approach by Ke Ma \etal \cite{AttrLayout2Im}, we used the official code\footnote{\url{https://github.com/ubc-vision/attribute-guided-image-generation-from-layout}} to train a model since no pre-trained checkpoint was available.
We trained the 128$\times$128 model using the default settings except we used a higher batch size of 12 (instead of 8) for 900k iterations which took roughly 20 days on a single NVIDIA A100 GPU.

\section{Evaluation Details}
The results for Layout2Im are as reported in \cite{LostGAN}.
For a fair comparison, we evaluate all other models ourselves.
To obtain the results for the baselines LostGANv1 and LostGANv2, we used the pre-trained models available at the official code repository\footnote{\url{https://github.com/WillSuen/LostGANs}}.
Our results are slightly different compared to the scores reported in the original publications \cite{LostGAN,LostGANv2}, and here we explain how we obtained our numbers.

\myparagraph{Image Format}
In contrast to the official codebase, in which the images are saved as JPG files, we save the images in PNG format.
Using JPG leads to artifacts due to compression and subsampling, which can act like a blur.
When computing the FID against real images that were passed through the dataloader and saved as JPG, the generated and real images become more similar due to the ``blur'' effect from using JPG.
This difference can lead to better scores and unfair comparisons which is why we re-evaluated all models.
Since we save the images in PNG format, the fine-grained details of real images are preserved thereby allowing a more robust comparison between real and generated images, which leads to worse FID scores.
Due to the same reason, we also report slightly higher IS scores obtained for real images.
The problem of different image formats has also been observed in \cite{casanova2020generating}.

\myparagraph{Data Splits}
Another difference stems from the choice of data splits used to compute the final scores.
To get the upper bound FID and SceneFID for real images, we compare real validation against real testing images (and object crops, respectively).
Similarly, we compute the FID and SceneFID for the models by comparing images generated from the validation set against real images from the testing set.
This is different to the procedure in \cite{LostGAN}, where images generated from validation layouts are compared against real validation images.
Since we aim to get an estimate of the real image distribution, using the same split for generation and comparison might overlook the diversity of real images, leading to a biased score.
For the same reason, we report the IS and SceneIS on images generated from layouts in the testing set.
Similarly, the DS is computed on two sets of images generated from the testing set using different random seeds.

\begin{figure*}[t]
\centering
\includegraphics[width=1.0\linewidth]{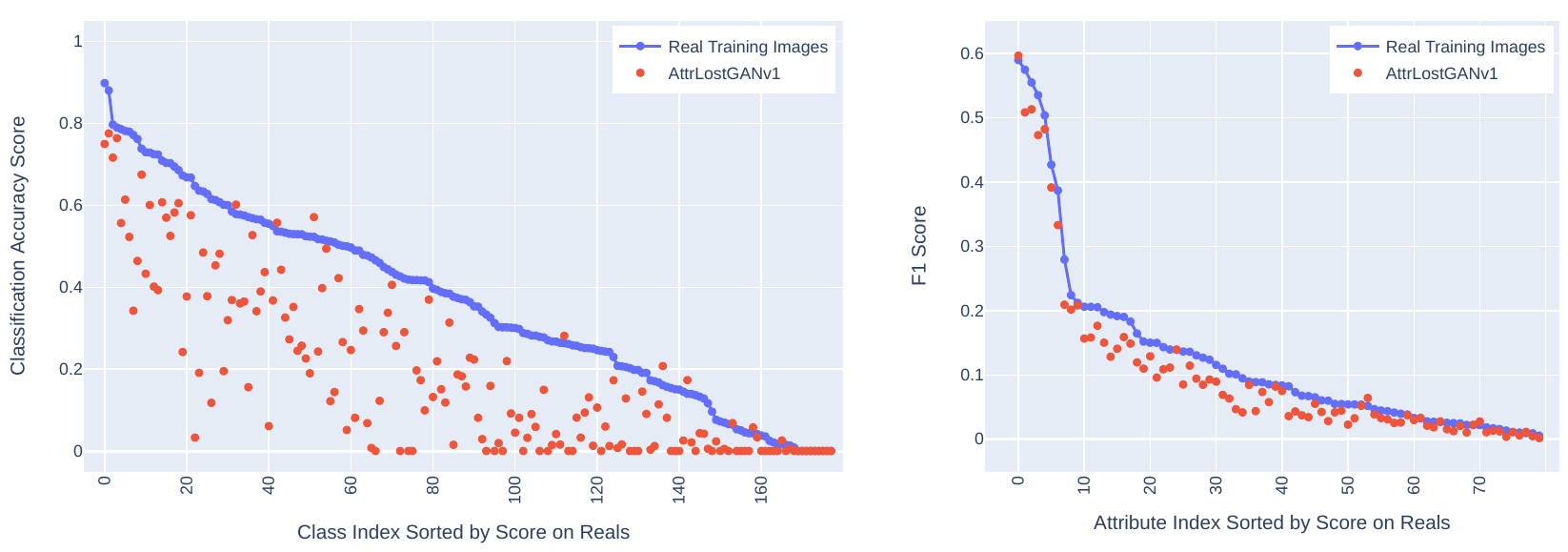}
   \caption{Comparison of per-class accuracy (left), and per-attribute F1 score of real image data (blue) vs. our AttrLostGANv1 (red) on 128$\times$128 resolution.
   The CAS score automatically surfaces challenging to model classes \cite{CAS}, and our proposed Attr-F1 provides insights into which attributes are not well-captured by the generated data distribution when compared to real images.
   While achieving good scores on most of the classes and attributes, there are many others that perform far below the performance achieved by real data which supports the intuition that current methods often struggle to generate recognizable objects in complex scene settings.
   }
\label{fig:cas_f1}
\end{figure*}

\myparagraph{CAS}
The intuition behind CAS \cite{CAS} is to ``train on fake, test on real'', which allows evaluating a long-standing goal of generative modelling, namely, whether one can replace real with synthesized images to train discriminative models.
To compute the scores, we first fine-tune an ImageNet pre-trained ResNet101 for each method for 20 epochs on generated data from the training layouts.
We then report the test accuracy on real data of the checkpoint with the lowest validation loss on real data.
This is different to the procedure in \cite{LostGAN}, where the classifier was both trained and tested on the same split using images generated from validation layouts, and tested on real validation images.
Furthermore, the CAS used in our work and \cite{LostGAN,LostGANv2} is different from the classification accuracy as used in \cite{Layout2Im,AttrLayout2Im}, which measures the accuracy by training on real and testing on generated images and hence might entirely disregard the diversity of generated images \cite{LostGANv2}.
Furthermore, CAS can reveal interesting failure cases, \eg, it can expose classes that are difficult to model \cite{CAS} (see next section).
Another benefit of CAS is that models cannot overfit the metric during training, \eg, when employing an object classification loss as in \cite{AttrLayout2Im}.

\myparagraph{Attr-F1}
Similar to the procedure described above, we proposed Attr-F1 to assess the attribute quality of the models.
To compute the scores, we first fine-tune an ImageNet pre-trained ResNet101 to perform multi-label classification on the attributes.
For each method, we fine-tune a classifier for 20 epochs on generated data from the training layouts using a weighted binary cross-entropy loss, and report the micro F1 score on real test data using the checkpoint with the lowest validation loss on real data.

\section{Analysis of Modelling Capacity}
Generating high-resolution images of complex scenes like the ones provided in Visual Genome \cite{VisualGenome} is challenging and current models still struggle.
The dataset we used contains 179 different object classes, and objects can have multiple attributes from a set of 80.
This leads to questions regarding the limitations of a generative method such as how many object classes and attributes are captured by the model, and which are easy and hard to model, respectively.
Here, we aim to provide insights into these questions in the setting of generating complex scenes of multiple objects by breaking down the performance on individual object classes and attributes.
Such a per-class, and per-attribute breakdown can uncover model deficiencies \cite{CAS} and allows more fine-grained comparisons with future models.
Our analysis sheds light upon how many different object classes and attributes a state-of-the-art layout-to-image model can capture, and with which the model struggles on the challenging Visual Genome dataset when compared to real images using a downstream discriminative task.
However, note that the image resolution is 128$\times$128 and hence many objects are very small and unrecognizable even for real images when viewed independently of the rest of the image.

\myparagraph{Object Classes}
The CAS \cite{CAS} can automatically surface object classes that are not well-captured by the generative model.
In \autoref{fig:cas_f1} (left), we plot the per-class accuracy of real image data against generated image data by our AttrLostGANv1.
As can be seen, our model is able to generate recognizable objects for most of the classes.
However, there are many classes for which the performance of generated data is far below the performance achieved by real data, which supports the intuition that current models still struggle with complex datasets and that many objects are not recognizable in the generated images.
The top five classes on generated data are ``sheep'', ``light'', ``mountain'', ``window'', and ``tree''.
Similar to the observation in \cite{CAS} we find six classes (when thresholding at 10\%) namely ``building'', ``pole'', ``arm'', ``logo'', ``stone'', and ``engine'', for which the generated data achieves a slightly higher accuracy.
The worst five classes as judged by the difference to the score of real image data are ``wood'', ``part'', ``bus'', ``giraffe'', and ``handle''.

\myparagraph{Attributes}
Inspired by CAS, we proposed Attr-F1 to evaluate the ability to generate objects with specified attributes.
While we report the micro F1 score in the main results, we can also break down the performance on a per-attribute level (similar to the object classes).
In \autoref{fig:cas_f1} (right), we plot the per-attribute F1 score of real image data against generated image data by our AttrLostGANv1.
Overall, the performance of generated data follows the performance of real data and does not vary as drastically as observed on object classes.
As can be seen, there are a few attributes that achieve relatively high scores as compared to most of the other attributes.
The top five attributes on generated data all refer to colors namely ``green'', ``blue'', ``white'', ``red'', and ``black'' which are both easy to generate and easy to detect.
We also observe three attributes (when thresholding at 5\%) namely ``green'', ``stone'', and ``calm'' which even achieve a slightly higher score than real data.
On the other hand, we can observe that attributes which are in general much more difficult to model (but also to classify) often refer to material aspects (\eg ``plastic'', ``leather'', ``dirty''), shape (\eg ``rectangular'', ``thick'', ``light''), and state (\eg ``walking'', ``playing'', ``open'', ``closed'', ``empty'').

\section{More Visual Results}
\begin{figure*}[th]
\centering
\includegraphics[width=1.0\linewidth]{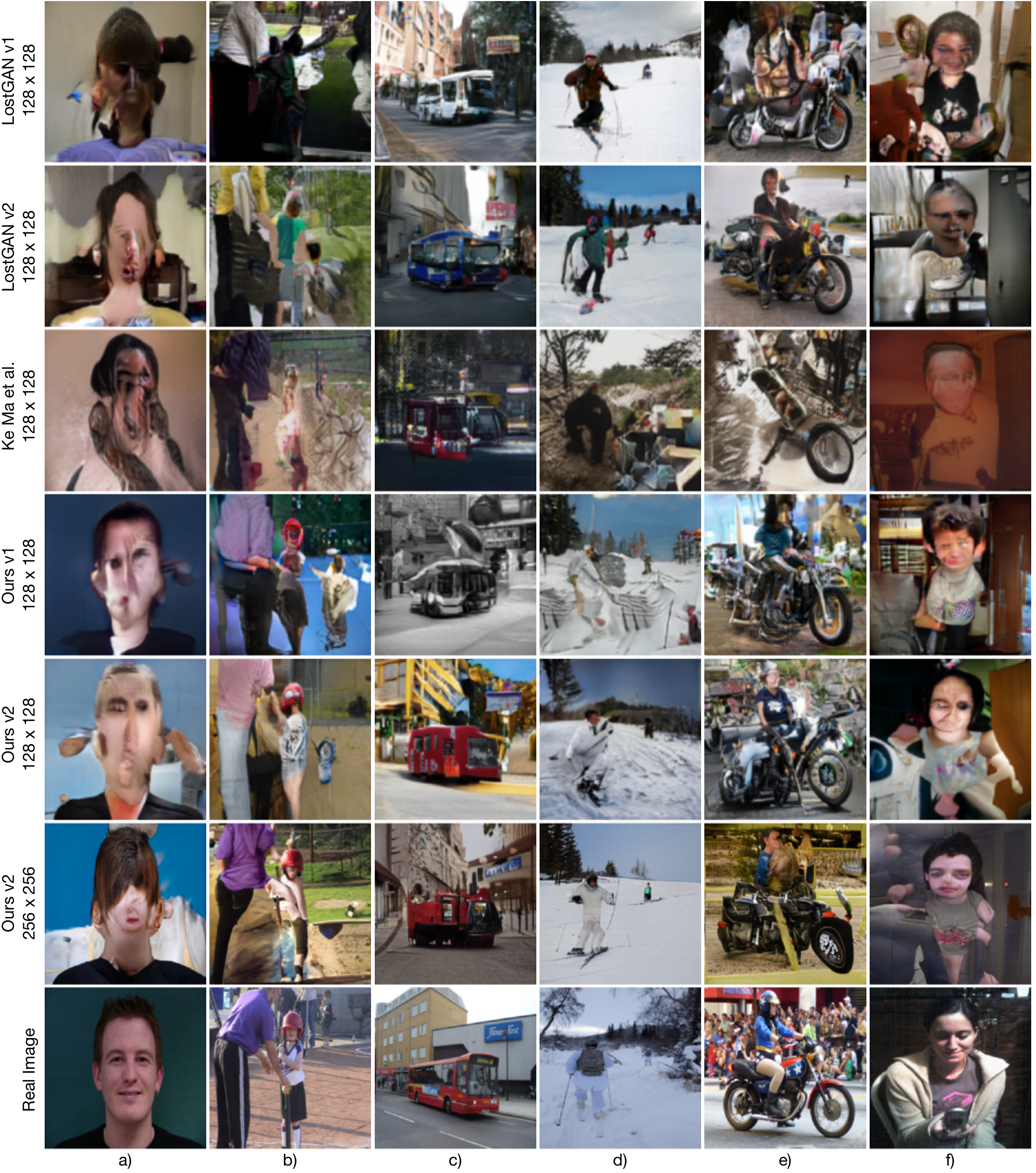}
   \caption{Comparison between methods.
   Our methods produce results that are visually more appealing and closer to the real images in terms of attribute specification.
   For example, note the following annotations as given in the original layout:
   a) shirt [black], wall [blue], b) shirt [purple], helmet [red], c) bus [red], sign [blue], d) coat [white], pant [white], e) jacket [blue], f) shirt [grey], logo [pink, purple].}
\label{fig:comparison}
\end{figure*}

\begin{figure*}
\centering
\includegraphics[width=1.0\linewidth]{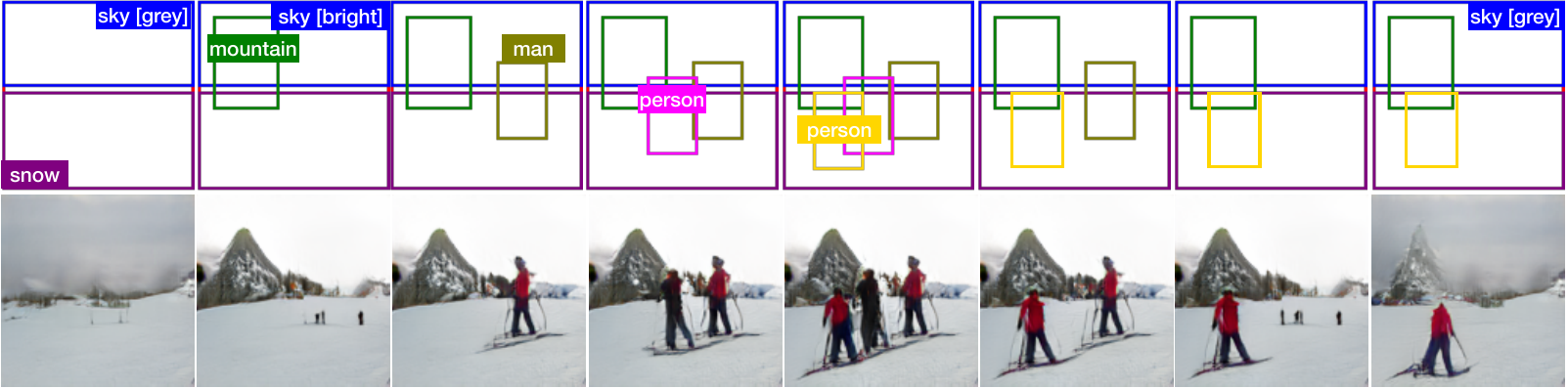}
   \caption{Example of iteratively adding and removing objects in a different order from the layout without affecting the overall image.
   \textit{From left to right:} sky [grey $\rightarrow$ bright], add mountain; add man; add person 1; add person 2; remove person 1; remove man; sky [bright $\rightarrow$ grey].}
\label{fig:add_and_remove}

\vspace{1em}
\includegraphics[width=1.0\linewidth]{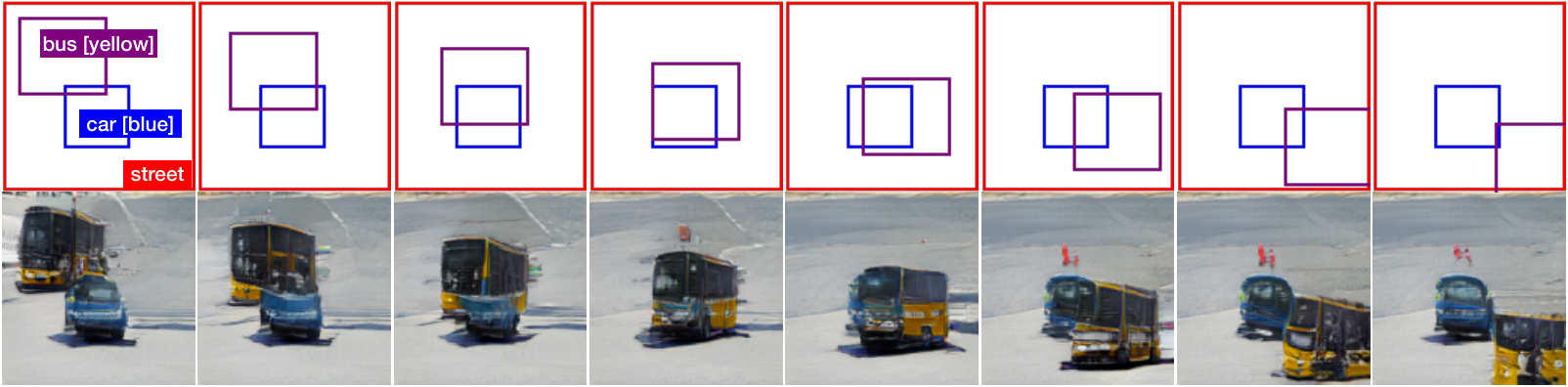}
   \caption{Example of diagonally shifting one object showing that nearby and overlapping objects influence each other's appearance.}
\label{fig:move_vg}

\vspace{1em}
\includegraphics[width=1.0\linewidth]{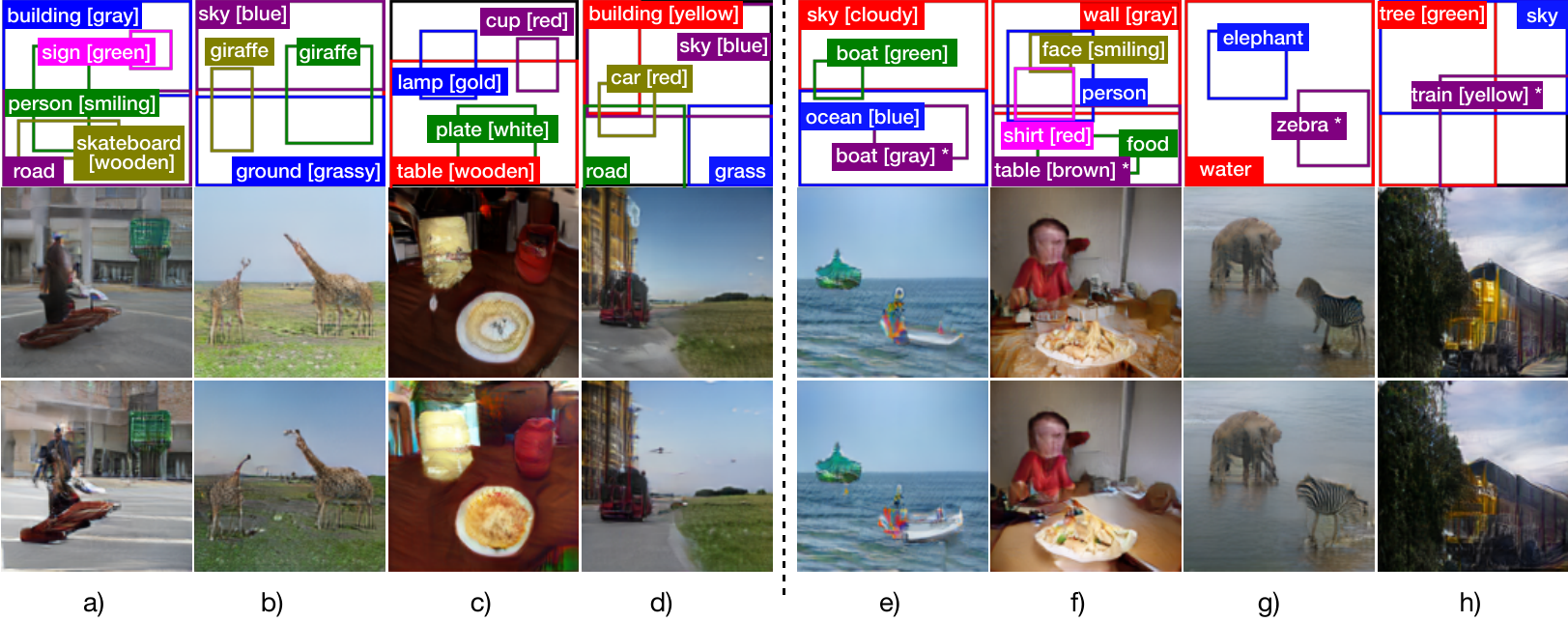}
   \caption{Our model allows to generate diverse images even if attributes are specified and we can control global vs.\ local appearance changes.
   In a) - d), we re-sample the image latent code $z_\textrm{img}$ and observe global appearance changes such lighting conditions.
   In e) - h), we re-sample the latent code for the object marked with * in the layout and observe local appearance changes.}
\label{fig:noise}
\end{figure*}

In \autoref{fig:comparison} we compare generated images between all methods using test layouts.
Due to the missing ability to specify attributes, the LostGAN \cite{LostGAN,LostGANv2} baselines produce random colors for the objects.
While \cite{AttrLayout2Im} sometimes correctly produces the attribute specifications, the overall image and objects are far from realistic.
Our methods produce the most realistic images which can be seen by comparing individual objects from real and generated images.
The objects have fine-grained details, and better reflect the input specifications.
Comparing with real ground-truth images, we can see that the images generated by our model are in general of higher quality and closer to real images in terms of attribute specification.
Visual Genome \cite{VisualGenome} contains many gray-scale images, and interestingly, we observed that the models sometimes produce gray-scale images even if color attributes are specified (see third row of AttrLostGANv1).
In \autoref{fig:add_and_remove}, we show that we can add and remove objects in a different order without affecting the overall image and other objects during reconfiguration.
In \autoref{fig:move_vg}, we diagonally move one object showing that nearby and overlapping objects influence each other's appearance which can be a desirable feature when modelling interactions between objects in more complex settings.
In \autoref{fig:noise}, we demonstrate diverse image synthesis by re-sampling the image and individual object latent codes.
This allows users to control the global and local appearance separately, even if attributes are specified.

\end{document}